\documentclass[runningheads]{llncs}
\def\BibTeX{{\rm B\kern-.05em{\sc i\kern-.025em b}\kern-.08em
    T\kern-.1667em\lower.7ex\hbox{E}\kern-.125emX}}
\usepackage{graphicx}
\usepackage{mathtools}
\usepackage{lipsum} 
\usepackage[numbers]{natbib}

\usepackage{booktabs}
\usepackage[ruled]{algorithm2e}
\usepackage{url}
\usepackage{color}
\usepackage{multirow}
\usepackage{enumitem}
\usepackage{array}
\usepackage{siunitx}
\usepackage[labelfont={bf}]{caption}
\usepackage{balance}
\usepackage{array} 
\usepackage{todonotes}
\usepackage{subfig}
\usepackage[nolist,nohyperlinks]{acronym}
\usepackage{footnote}
\usepackage[perpage]{footmisc}
\usepackage[english]{babel}
\usepackage[autostyle]{csquotes}
\makesavenoteenv{tabular}

\acrodef{CNN}{Convolutional Neural Network}
\acrodef{AI}{Artificial Intelligence} 
\acrodef{ML}{Machine Learning}
\acrodef{DL}{Deep Learning}
\acrodef{DNN}{Deep Neural Network}
\acrodef{SVD}{singular value decomposition}

\begin{document}
\title{LightLayers: Parameter Efficient Dense and Convolutional Layers for Image Classification}


\author{Debesh Jha\inst{1,2}\and 
Anis Yazidi \inst{3} \and 
Michael A. Riegler\inst{1}\and \\
Dag Johansen\inst{2}\and
H\aa vard D. Johansen\inst{2}\and 
P\aa l Halvorsen\inst{1,3}
}

\authorrunning{Jha et al.}
\titlerunning{LightLayers}

\institute{SimulaMet, Norway \and UIT The Arctic University of Norway \and Oslo Metropolitan University, Norway\\
\email{debesh@simula.no}
}

\maketitle            
\begin{abstract}
Deep Neural Networks (DNNs) have become the de-facto standard in computer vision, as well as in many other pattern recognition tasks. A key drawback of DNNs is that the training phase can be very computationally expensive. Organizations or individuals that cannot afford purchasing state-of-the-art hardware or tapping into cloud hosted infrastructures may face a long  waiting time before the training completes or might not be able to train a model at all.  Investigating novel ways to reduce the training time could be a potential solution to alleviate this drawback, and thus enabling more rapid development of new algorithms and models. 
In this paper, we propose LightLayers, a method for reducing the number of trainable parameters in deep neural networks (DNN). The proposed LightLayers consists of LightDense and LightConv2D layer that are as efficient as regular Conv2D and Dense layers, but uses less parameters. We resort to Matrix Factorization to reduce the complexity of the DNN models resulting into lightweight DNN models  that require less computational power, without much loss in the accuracy.  We have tested LightLayers on MNIST, Fashion MNIST, CIFAR 10, and CIFAR 100 datasets. Promising results are obtained for MNIST, Fashion MNIST, CIFAR-10 datasets whereas CIFAR 100 shows acceptable performance by using fewer parameters.

\vspace{-5mm}
\end{abstract}

\vspace{-5mm}
\section{Introduction}
\label{sec:introduction}
Deep learning techniques have revolutionized the field of \ac{ML} and gained immense research attention during the last decade. Deep neural network provides state-of-the-art solution in several domains such as image recognition, speech recognition, and text processing~\cite{novikov2015tensorizing}. One of the most popular techniques within deep learning is \acf{CNN}, which possesses a structure that is well-suitable specially for image and video processing. A \ac{CNN}~\cite{lecun1998gradient} comprises a convolution layer and dense layer. CNN has emerged as powerful techniques for solving many classification~\cite{krizhevsky2012imagenet} and regression~\cite{kleinbaum2002logistic} tasks. Additionally, \ac{CNN} has produced promising results in various applications areas, including in the medical domain, with applicability in diabetic retinopathy prediction~\cite{arcadu2019deep}, endoscopic disease detection~\cite{thambawita2020extensive}, and breast cancer detection~\cite{mckinney2020international}. 

Recently, developing deeper and larger architectures has been a common trend in the development of state-of-the-art methods~\cite{brown2020language}. Most of the time, we can observe that deeper networks especially with large and complex datasets lead to better performance. One of the major drawbacks of \acp{CNN} are that they  often require an immense amount of training time compared to other classical \ac{ML} algorithms. Hyperparameter optimization for fine-tuning the model is another challenging task that increases dramatically the overall training time to achieve optimum results from any model. \ac{CNN} models often require powerful Graphical Processing Units (GPUs) for training, which can span over days, weeks, and even months, with no guarantee that the model will produce satisfactory results. A long training process also consumes a lot of energy and is not considered environmental friendly.
Furthermore, long training is demanding in terms of resources as a large amount of memory is required which renders it difficult to deploy it into low-power devices~\cite{kim2015compression}. The requirements for the expensive hardware and high training time complicate the use of models with large number of trainable parameters to be deployed it into portable devices or conventional desktops~\cite{novikov2015tensorizing}. 

A potential way to address these issues is the introduction of lightweight models. A lightweight model can potentially be built by reducing the number of trainable parameters within the layers. In an effort towards reducing the training time and complexity  of \ac{CNN} models, we propose LightLayers, which is a combination of LightDense and LightConv2D layers, that focuses on \acp{CNN} and more particularly on creating both a lightweight convolutional layer and a lightweight dense layer that are both easy to train. Lightweight \ac{CNN} models are computationally cheap and can be deployed various applications for carrying out online estimation. Therefore, the main goal of the paper is to present a general model to reduce the number of parameters in a \ac{CNN} model so that it can be used in various image processing or other applicable tasks in the future.

The main contributions of the paper are:
\begin{itemize}
    \item LightLayers, a combination of LightConv2D and LightDense layers, is proposed.  Both layers are based on matrix decomposition for reducing the number of trainable parameters of the layers. 

    \item We have investigated and tested the proposed model with four different publicly available datasets: MNIST~\cite{lecun1998gradient}, Fashion MNIST~\cite{xiao2017fashion}, CIFAR10~\cite{krizhevsky2009learning}, CIFAR100~\cite{krizhevsky2009learning}, and showed that the proposed method is competitive in terms of both accuracy and efficiency when the number of training parameters used are taken into consideration. 

    \item We experimentally show that good accuracy can be achieved by using a relatively small number of trainable parameters with MNIST, Fashion MNIST, and CIFAR 10 dataset. Moreover, we found there was a significant reduction in the number of trainable parameters as compared to Conv2D.

\end{itemize}


\section{Related Work}
\label{sec:relatedwork}

In the context of reducing the cost of network model training, several approaches have been presented. For example, Xue et al.~\cite{xue2013restructuring} presented a \acf{DNN} technique for reducing the model size while maintaining the accuracy. For achieving this goal, they  used \ac{SVD} on the weight matrix in \ac{DNN}, and reconstructed the model based on inherent sparseness of the original matrices. The application of \ac{DNN} for mobile applications has become increasingly popular. The computational and storage limitation should be taken into account while deploying \ac{DNN} into such devices.

To address this need, Li et al.~\cite{li2014learning} proposed two techniques for effectively learning from \acp{DNN} with a smaller number of hidden nodes and smaller number of senones set. The details about both the techniques can be found in the literature~\cite{li2014learning}. Similarly, Xue et al.~\cite{xue2014singular} introduced two \ac{SVD} based techniques to solve the issue related to \ac{DNN} personalization and adaptation. Garipov et al.~\cite{garipov2016ultimate} developed a tensor factorization framework for compressing fully-connected layer. The focus of their work was to compress convolutional layers which would potentially excel in image recognition tasks by reducing the memory complexity and high computational cost. Later, Kim et al.~\cite{kim2017kernel} proposed energy-efficient kernel decomposition architecture for binary-weight \acp{CNN}.

Ding et al.~\cite{ding2017circnn} proposed CIRCNN, an approach for representing the weights and processing neural networks by the use of block-circulant matrices. CIRCNN utilizes Fast Fourier Transform based fast multiplication operation which simultaneously reduces the computational and storage complexity causing negligible loss in accuracy. Chai et al.~\cite{wu2018prodsumnet} proposed a model for reducing the parameters in deep neural networks via product-of-sums matrix decomposition. They obtained good accuracy on the MNIST and Fashion MNIST datasets with a smaller number of trainable parameters. Another similar work is by Agrawal et al.~\cite{agarwal2019lightweight}, where they designed a lightweight deep learning model for human activity recognition that is sufficiently computationally efficient to be deployed in edge devices. For more recent works on matrix and tensor decomposition, we refer the reader to~\cite{denton2014exploiting,lebedev2014speeding}. 

Kim et al.~\cite{kim2015compression} proposed a method for compressing \ac{CNN} to be deployed into a mobile application. Mariet et al.~\cite{mariet2015diversity} proposed another efficient neural network architecture that reduces the size of neural network without hurting the overall performance. Novikov et al.~\cite{novikov2015tensorizing} converted dense weight matrices of fully-connected layers to Tensor Train~\cite{oseledets2011tensor} format such that the number of parameters are reduced by huge factor by preserving the expressive power of the layer.

Lightweighted networks have gained attention in computer vision (for instance, in the area of real-time image segmentation~\cite{jiang2020lrnnet,paszke2016enet,wang2019lednet,yu2018bisenet}). Real-time applications are growing because the lightweight models can be an efficient solution for resource constraints and mobile devices. Only a lightweight model demands lower memory that leads to a lower computation and faster speed. Therefore, developing a lightweight model can be a good idea for achieving real-time solutions, and it can also be used for other applications too. \\


The above studies show that there is great potential for lightweight networks for computer-vision tasks. With large amounts of training data, it is likely that a model with huge numbers of trainable parameters will outperform the smaller models---if one can afford the high training costs and resource demands at inference time. However, there is a need for models with low-cost computational power and small memory footprints~\cite{kim2015compression}, especially for mobile applications~\cite{kim2015compression} and portable devices. 
In this repsect, we propose LightLayers 
that are based on the concept of matrix decomposition. LightLayers uses fewer trainable parameters and shows the state-of-the-art tradeoff between parameter size and accuracy. 


\section{Methodology}
\label{sec:Methodology}
\begin{figure} [!t]
    \centering
    \includegraphics [width = 0.6\columnwidth] {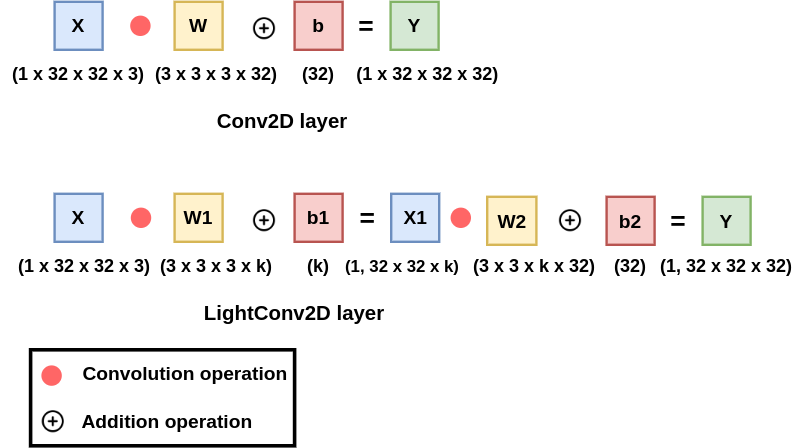}
    \caption{Comparative diagram of Conv2D layer and LightConv2D layer}
    \label{figure1}
\end{figure}

\begin{figure} [!t]
    \centering
    \includegraphics [ width = 0.6\columnwidth] {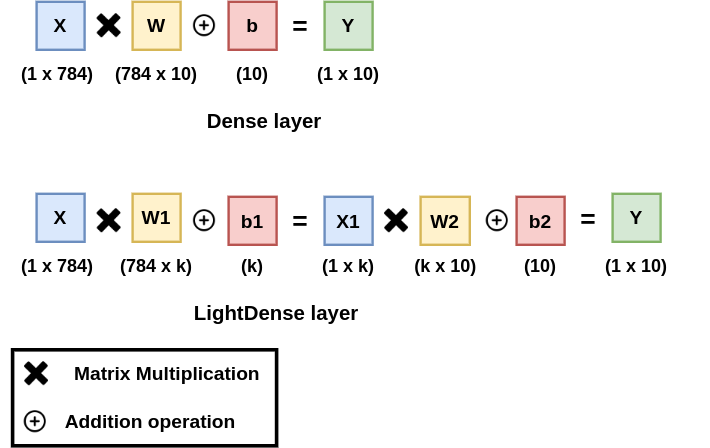}
    \caption{Comparative diagram of Dense layer and LightDense layer}
    \label{figure2}
\end{figure}

In this section, we introduce the proposed layers. Figure~\ref{figure1} shows the comparison of a Conv2D and a LightConv2D layer. In the LightConv2D layers, we decompose the weight matrix $W$ into $W1$ and $W2$ on the basis of hyperparameter $k$, which leads to a reduction of the total number of trainable parameters in the network. We follow the same strategy for the LightDense layer. The block diagram of the LightDense layer is shown in Figure~\ref{figure2}.

\begin{figure} [!t]
    \centering
    \includegraphics[height=12cm, angle=90]{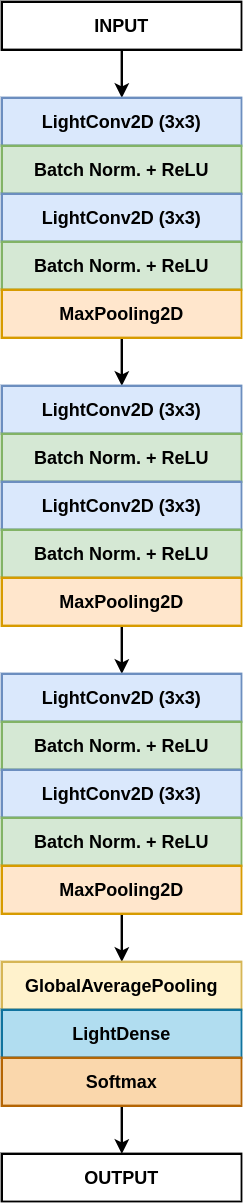}
    \caption{Block diagram of the architecture used for comparison of the proposed Lightlayers with regular convolution and dense layers. In the case of regular layers, we use regular convolution and dense layers instead of Lightlayers.}
    \label{figure3}
    \vspace{-5mm}
\end{figure}

The main objective of building the model is to compare our LightLayers (i.e., the combination of LightConv2D and LightDense layers) with the conventional Conv2D and SeparableConv2D layers.  For comparing the performance of the various layers, we have built a simple model from scratch.
The block diagram of the proposed model is shown in Figure~\ref{figure3}. We used the same hyperparameters and setting for all the experiments. For the LightLayers experiments, we used LightConv2D and LightDense layers (see Figure~\ref{figure3}). For the other experiments, we replaced LightConv2D with Conv2D or SeparableConv2D and LightDense with a regular Dense layer.

The model architecture used for experimentation (see Figure~\ref{figure3}) comprises  two $3\times 3$ convolution layers, each followed by a batch-normalization and ReLU non-linearity as the activation function. We have introduced $2\times2$ max-pooling, which reduces the spatial dimension of the feature map. We have used three similar blocks of layers in the model followed by the GlobalAveragePooling, LightDense layers with $k$ = $8$, and a softmax activation function for classifying the input image.

\vspace{-5mm}

\subsection{Description of convolution layers}
\subsubsection{Conv2D}
A convolution layer is the most common layer used in any computer vision task and is applied extensively. This layer uses a multidimensional kernel as the weight, which is used to perform convolution operation on the input to produce an output. If the bias is used, then a $1D$  vector is added to the output. Finally, the activation is applied to introduce the non-linearity into the neural network. In this paper, we worked on a $2D$ convolution layer, which uses a $4D$ tensor as the weight.
\vspace{-1mm}
\begin{equation} {\label{eq:1}}
Output = Activation((Input \otimes Weight) + Bias) 
\end{equation}
In the above equation, $\otimes$ represents the convolution operation, and weight represents the kernel.

\subsubsection{Dense layer}
\vspace{-5mm}
A dense layer is the regular, deeply connected neural-network layer. It is the most common and frequently used layer. It is also known as a fully-connected layer as each neuron receives input from the previous layer. 
\begin{equation} {\label{eq:2}}
Output = Activation((Input \oplus Weight) + Bias)    
\end{equation}
In the above equation, $\oplus$ represents the matrix multiplication instead of convolution operation as above.

\subsubsection{Separable Conv2D}
\vspace{-5mm}
Separable convolution, also known as depth-wise convolution, is used in our experiment. We use depth-wise separable 2D convolution to compare the performance of our model. It first applies a depth-wise spatial convolution, i.e., performing convolution operation on each input channel independently. After that, it is followed by a point-wise convolution, i.e., a $1\times1$ convolution. Point-wise, convolution controls the number of filters in the output feature maps.


\section{Experimental Setup}
\label{sec:Experiments}
For the experiments, we use the same number of layers, filters, filter sizes, and activation functions in every model for the individual dataset. We have modified the existing Dense and Conv2D layer in such a way that the number of trainable parameters decreases with some decrease in the accuracy of the model.  In particular, we use three types of layers for this experiment, i.e., Conv2D, SeparableConv2D, and LightLayers. First, we run the model using Conv2D layers. The Conv2D layer is replaced by SeperableConv2D and run again. Again, we replace SeperableConv2D with the LightLayers and run the model.

In the modified layers, we introduced the hyperparameter $k$ to control the number of trainable parameters in the LightDense and LightConv2D layer. In the LightDense layer, we set $k$ to $8$. In the LightConv2D layer, $k$ varies between $1$ to $6$, and more could be set depending on the requirement.   The values of the $k$  are chosen empirically. We only replace the Conv2D layer with the LightConv2D layer and Dense layer with the LightDense layer of the proposed lightweight model. The rest of the network architecture remains the same. 
\vspace{-5mm}
\subsection{Implementation Details}
We have implemented proposed layers using the Keras framework~\cite{chollet2015keras} and TensorFlow 2.2~\cite{abadi2016tensorflow} as backend. The implementation can be found at GitHub\footnote{\url{https://github.com/DebeshJha/LightLayers}}. We performed all the experiments on an NVIDIA GEFORCE GTX 1080 system, which has 2560 NVIDIA CUDA Cores with 8 GB GDDR5X memory. The system was running on Ubuntu 18.04.3 LTS. We used a batch size of 64. All the experiments were run, keeping all the hyperparameters (i.e., learning rate, optimizer, batch size, number of filters, and filter size) the same. We have trained all the models for 20 epochs. After each convolution layer, batch normalization is used, which is activated by the Rectified linear unit (ReLU).


\subsection{Datasets}
To evaluate LightConv2D layer and LightDense layer, we have performed experiments using various datasets.
\subsubsection{MNIST Database}
\label{sec:MNIST}
\vspace{-4mm}
Modified National Institute of Standards and Technology (MNIST)~\cite{lecun1998gradient}  is the primary dataset for computer vision tasks introduced by LeCun et al. in 1998. MNIST comprises $10$ classes of handwritten digits with $60,000$ training and $10,000$ testing images. The resolution of the images in the MNIST dataset is $28\times 28$.  There is a huge recent advancement in \ac{ML} and \ac{DL} algorithms. However, the MNIST remains a common choice for learners and beginners. The reason is that it is easy to deploy, test, and compare an algorithm on a publicly available dataset. The dataset can be downloaded from \url{http://yann.lecun.com/exdb/mnist/}.
 
 \vspace{-3mm}
\subsubsection{Fashion MNIST Database}
Fashion MNIST~\cite{xiao2017fashion} is a $10$ class of $70,000$ grayscale images of size $28\times 28$. Han et al. released a novel image dataset that could be used for benchmarking \ac{ML} algorithms. Their goal was to replace the MNIST database with a new database. The images of the Fashion MNIST database are more challenging as compared to the MNIST database. It contains natural images such as t-shirt/top, trouser, pullover, dress, coat, sandal, shirt, sneaker, bag, and ankle boot. The database can be downloaded from \url{https://github.com/zalandoresearch/fashion-mnist}. 
\vspace{-3mm}
\subsubsection{CIFAR-10 Database}
CIFAR-10~\cite{krizhevsky2009learning} is a commonly established dataset for computer-vision tasks. It is especially used for object recognition tasks. CIFAR-10 contains $60,000$ color images of size $32\times32$. It also has $10$ classes of images. Each class contains $6000$ images per class. The classes contain datasets of cars, birds, cats, deer, dogs, horses, and trucks. The dataset can be downloaded from \url{https://www.cs.toronto.edu/~kriz/cifar.html}.
\vspace{-3mm}
\subsubsection{CIFAR-100 Database}
CIFAR-100~\cite{krizhevsky2009learning} is also collected by the team of Alex Krizhevsky, Vinod Nair, and Geoffrey Hinton. This database is similar to the previous CIFAR-10 database. The $100$ classes of the database consist of images such as beaver, dolphin, flatfish, roses, clock, computer keyboard, bee, forest, baby, pine, tank, etc. Each class of the database contains $600$ images each. This dataset contains $500$ training examples and $100$ testing examples per class. The dataset can be found on the same webpage as CIFAR-10.

\section{Results}
\label{sec:Results}
In this section, we present and compare the experimental results of the Conv2D, SeperableConv2D, and LightLayers models on the MNIST, Fashion MNIST, CIFAR-10, and CIFAR 100 datasets. Table~\ref{table1} shows the summary of result comparison of Conv2D, SeperableConv2D, and LightLayers on MNIST dataset. Based on Conv2D and SeperableConv2D, we propose Layers and show improvement over both layers. The concept of LightLayers are based on weight matrix decomposition. This is the main motivation behind comparison of the proposed layers with Conv2D and SeperableConv2D.  

The hyperparameters used are described in the caption of the Table~\ref{table1}. We can see that the result of the proposed LightLayers is comparable to that of Conv2D and SeperableConv2D in terms of test accuracy.  When we compare the LightLayers with Conv2D, in terms of the number of parameters used, it uses only \( \frac{1}{3} \) of parameters of Conv2D, which is more efficient with only $1\%$ drop in terms of test accuracy. LightLayers with hyperparameter $k=3$ achieves the highest test accuracy. However, for the other values of $k$ as well there is only minimal variation in test accuracy. 

\begin{table}[!t]
\def\arraystretch{1.2}
    \setlength\tabcolsep{5pt}
    \par\bigskip
    \centering
\caption{Results on \textbf{MNIST} test dataset (Number of epochs = 10, Batch size = 64, Learning rate = $1e-3$, Number of filters = [$8, 16, 32$]).}  
\label{table1}
\begin{tabular}{|l|c|c|c|}
\hline
\textbf{Method} & \textbf{Parameters} & \textbf{Test Accuracy} & \textbf{Test Loss} \\ \hline
Conv2D & 18,818 & 0.9887 & 0.018  \\ \hline
SeparableConv2D & 3,611 & 0.9338 & 0.2433  \\ \hline
LightLayers ($K$ = $1$) & 2,649 & 0.9418 & 0.1327 \\ \hline
LightLayers ($K$ = $2$) & 4,392 & 0.9749 & 0.0554 \\ \hline
\textbf{LightLayers} ($K$ = $3$) & \textbf{6,135} & \textbf{0.9775} & \textbf{0.0513} \\ \hline
LightLayers ($K$ = $4$) & 7,878 & 0.9720 & 0.0704 \\ \hline
\end{tabular}
\end{table}

\begin{table}[!t]
\def\arraystretch{1.2}
    \setlength\tabcolsep{5pt}
    \par\bigskip
    \centering
\caption{Results on \textbf{Fashion MNIST} test dataset (Number of epochs = 10, Batch size = 64, Learning rate = $1e-3$, Number of filters = [$8, 16, 32$]).}
\label{table2}
\begin{tabular}{|l|c|c|c|}
\hline
\textbf{Method} & \textbf{Parameters} & \textbf{Test Accuracy} & \textbf{Test Loss} \\ \hline
Conv2D & 18,818 & 0.9147 & 0.1468   \\ \hline
SeparableConv2D & 3,611 & 0.8725 & 0.3175 \\ \hline
LightLayers ($K$ = $1$) & 2,649 & 0.789 & 0.6752  \\ \hline
LightLayers ($K$ = $2$) & 4,392 & 0.8452 & 0.4247 \\ \hline
LightLayers ($K$ = $3$) & 6,135 & 0.8695 & 0.3708 \\ \hline
LightLayers ($K$ = $4$) & 7,878 & 0.8623 & 0.6184 \\ \hline
\textbf{LightLayers} ($K$ = $5$) & \textbf{9,621} & \textbf{0.8820} & \textbf{0.2810} \\ \hline
LightLayers ($K$ = $6$) & 11,364 & 0.8733 & 0.3986 \\ \hline

\end{tabular}
\end{table}

Table~\ref{table2} shows the results for different layers for the model trained on the Fashion MNIST dataset. From the table, we can see that the proposed model (LightLayers) with hyperparameter $k=5$ uses only half of the parameters with around drop $3\%$ drop in terms of test accuracy with the Fashion MNIST dataset. However, when we compare the quantitative results with SeperableConv2D, our proposed LightLayers achieves better test accuracy with the trade-off in number of trainable parameters. 

\begin{table}[!t]
\def\arraystretch{1.2}
    \setlength\tabcolsep{5pt}
    \par\bigskip
    \centering
\caption{Evaluation results on test set of \textbf{CIFAR10} dataset (Number of epochs = 20, Batch size = 64, Learning rate = $1e-4$, Number of filters = [$8, 16, 32,64$]). The 'Params' in the bold represents total number of parameters.}
\label{table3}
\begin{tabular}{|l|c|c|c|}
\hline
\textbf{Method} & \textbf{\shortstack{Parameters}} & \textbf{Test Accuracy} & \textbf{Test Loss}\\ \hline
Conv2D & 76,794 & 0.6882 & 0.9701 \\ \hline
SeparableConv2D & 14,440 & 0.5953 & 1.3263 \\ \hline
LightLayers  ($K$ = $1$) & 5,937 & 0.3686 & 1.6723  \\ \hline
LightLayers ($K$ = $2$) & 9,592 & 0.4596 & 1.5372  \\ \hline
LightLayers ($K$ = $3$)& 13,247 & 0.4937 & 1.5287  \\ \hline
LightLayers ($K$ = $4$)& 16,902 & 0.5319 & 1.3214  \\ \hline
\textbf{LightLayers} ($K$ = $5$) & \textbf{20,557} & \textbf{0.5576} & \textbf{1.2122 } \\ \hline

\end{tabular}
\end{table}

\begin{table}[!t]
\def\arraystretch{1.2}
    \setlength\tabcolsep{5pt}
    \par\bigskip
    \centering
\caption{Evaluation on \textbf{CIFAR100} test set (Number of epochs = 20, Batch size = 64, Learning rate = $1e-4$, Number of filters = [$8, 16, 32,64$]).}
\label{table4}
\begin{tabular}{|l|c|c|c|}
\hline
\textbf{Method} & \textbf{Parameters} & \textbf{Test Accuracy} & \textbf{Test Loss} \\ \hline

Conv2D & 82,644 & 0.3262 & 2.6576  \\ \hline
SeparableConv2D & 20,290 & 0.2207 & 3.2108  \\ \hline
LightLayers ($K$ = $1$)  & 6,747 & 0.0275 & 4.2391 \\ \hline
LightLayers ($K$ = $2$)  & 10,402 & 0.0398 & 4.1836   \\ \hline
LightLayers ($K$ = $3$) & 14,057 & 0.0559 & 4.0304  \\ \hline
LightLayers ($K$ = $4$) & 17,712 & 0.0551 & 3.9978 \\ \hline
\textbf{LightLayers} ($K$ = $5$) & \textbf{21,367} & \textbf{0.0589} & \textbf{4.0009 } \\ \hline
\end{tabular}
\end{table}

Table~\ref{table3} shows the results on CIFAR 10 dataset. On this dataset as well, the proposed method is $3.75$ times computationally efficient in terms of parameters it uses. However, there is a drop in accuracy of around $13\%$. Nevertheless, for some tasks the efficiency can be more important than the reduced accuracy. 

Similarly, we have trained and tested the proposed model on the CIFAR 100 dataset, where the test accuracy of the proposed layers is much lower as compared to the Conv2D. This is obvious because CIFAR 100 consists of 100 classes of images that are difficult to generalize with such a small number of trainable parameters. However, the total number of parameters used is still around $4$ times less than that of Conv2D. The total  number of trainable parameter for Conv2D is 82,644, and for LightLayers, it is only 21,367. More details on the test accuracy and test loss can be found from Table~\ref{table4}.

From the experimental results, we can say that LightLayers has the following advantages:
\begin{itemize}
    \item It requires less trainable parameters than Conv2D which is an important factor to implement it in different applications where heavy trainable parameters could not be beneficial. 
    \item  Due to less parameters, the space taken by the weight file is smaller which makes it more suitable to devices where storage space is limited. 
\end{itemize}

\vspace{-5mm}
\section{Ablation Study}
\label{ablationstudy}
Let us consider that the input size is $784$, and the number of output features is $10$. Therefore, the weight matrix $W$ is $784\times10$ resulting in $7840$ trainable parameters. Now, in the LightDense layer, we decompose the weight matrix $W$ into two smaller matrix $W1$ and $W2$ of lower dimension using the \mbox{hyperparameter $k$}. 

Here, $W1 = [784, k]$ and $W2=[k, 10]$ values from the above example, the total number of trainable parameters in the LightDense layer becomes $786\times k$ + $k \times 10$. Now, if $k=1$, then trainable parameters are $796$, and if $k=2$ the number of trainable parameters becomes $1,588$, and so on.

Next, consider the weight decomposition in the LightConv2D layer. If the input is $32 \times 32 \times 3$, the number of filters is $32$, and the kernel size is $3 \times 3$, then the filters size becomes $3 \times 3 \times 3 \times 32$. This means that the total number of trainable parameters is $864$. Now, we will decompose the kernel $W$ into $W1$ and $W2$ using hyperparameter $k$. Here, $W1$ is $3 \times 3 \times 3 \times k$ and $W2$ is $3 \times 3 \times k \times 32$. If $k$ is $1$, then the total number of trainable parameters becomes $27 + 288$, which is equals to $315$.

From the ablation study, we see that the number of trainable parameters used is less in LightLayers  compared to the Conv2D and Dense layers.  Overall, we can argue that the proposed LightLayers approach has the potential to be a powerful solution to solve the problem of excessive parameter used by traditional deep-learning approaches. However, our LightLayers model needs further improvement for successfully implementing it on a larger datasets with high resolution images. We can conclude that further investigating  matrix weight decomposition is important and other similar studies are necessary to reach the goal of lightweight models in the near future. 

\vspace{-3mm}
\section{Conclusion}
\label{sec:Conclusion}
In this paper, we propose the  LightLayers model, which uses matrix decomposition to help to reduce the complexity of the deep learning network. With the extensive experiments, we observed that changing the value of hyperparameter $k$ yields a trade-off between model complexity in terms of the number of trainable parameters and performance. We compare the accuracy of the LightLayers model with Conv2D.  An extensive evaluation shows the tradeoffs in terms of parameter uses, accuracy and computation.  In the future, we want to train LightLayers on  different publicly available datasets. We also aim to develop efficient techniques for finding the optimal value of $k$ automatically. Further research will be required to find suitable algorithms and implementations that will scale this approach to a biomedical datasets.



\vspace{-4mm}

\bibliographystyle{splncsnat}
\bibliography{references.bib}
\end{document}